\documentclass[10pt, a4paper, onecolumn]{article}

\usepackage[numbers]{natbib}

\usepackage[T1]{fontenc}
\usepackage{lmodern}

\usepackage{natbib}
\usepackage[utf8]{inputenc} 
\usepackage{booktabs}       
\usepackage{amsfonts}       
\usepackage{nicefrac}       
\usepackage{microtype}      
\usepackage{xcolor}         
\usepackage{graphicx}
\usepackage{amsmath,amssymb,amsfonts}
\usepackage{bm}

\usepackage{url}
\usepackage{hyperref}

\usepackage{algorithm}
\usepackage{algorithmicx}
\usepackage{algpseudocode}

\usepackage{array,multirow,graphicx}

\usepackage{geometry}
\newgeometry{vmargin={15mm}, hmargin={12mm,17mm}} 

\title{The Gaze and Mouse Signal as additional Source for User Fingerprints in Browser Applications}

\author{
	Wolfgang Fuhl, Daniel Weber, Shahram Eivazi \\
	Department of Human Computer Interaction\\
	University Tübingen\\
	Tübingen, 72076 \\
	\texttt{wolfgang.fuhl@uni-tuebingen.de} \\
	\texttt{daniel.weber@uni-tuebingen.de} \\
	\texttt{shahram.eivazi@mnf.uni-tuebingen.de} \\
}

\begin{document}
	
	\maketitle
	
	\begin{abstract}
		In this work, we inspect different data sources for browser fingerprints. We show which disadvantages and limitations browser statistics have and how this can be avoided with other data sources. Since human visual behavior is a rich source of information and also contains person specific information, it is a valuable source for browser fingerprints. However, human gaze acquisition in the browser also has disadvantages, such as inaccuracies via webcam and the restriction that the user must first allow access to the camera. However, it is also known that the mouse movements and the human gaze correlate and therefore, the mouse movements can be used instead of the gaze signal. In our evaluation, we show the influence of all possible combinations of the three information sources for user recognition and describe our simple approach in detail. \\
		Link: \url{https://es-cloud.cs.uni-tuebingen.de/d/8e2ab8c3fdd444e1a135/?p=%2FThe%20Gaze%20and%20Mouse%20Signal%20as%20additional%20Source%20...&mode=list}
	\end{abstract}

	\section{\uppercase{Introduction}}
	User identification plays a crucial role in many industrial sectors. In its original form, it is used to protect data and access to networks or premises~\cite{lee2010online,choubey2013secure}. Today, there is a growing need for user identification, especially in the online environment, which includes both personalized advertising~\cite{tucker2014social} and product placement~\cite{shamdasani2001location,fossen2019measuring}, but also online banking~\cite{lee2010online} or external access to corporate networks~\cite{cole2011network}. For security-critical applications such as external access to company networks or online banking, user IDs and passwords have become widely accepted. When using security-critical functionalities, additional security prompts such as a generated PIN or SMS prompts are added. In online advertising, as well as product placement, companies try to identify a person without accessing critical personal data. This is guaranteed in the modern world by so-called cookies~\cite{juels2006cache} since those have to be activated by the user, stateless approaches only use  browser statistics~\cite{juels2006cache}. A disadvantage of this method is that the statistics can be used to identify a computer very effectively, but in the case of computers with multiple users, all of them are treated as the same person. For the password and user recognition procedure, there are also disadvantages. For example, if the identification and password is known by an attacker, the attacker can gain access. 
	
	In this work, we analyze new data sources, like the eye signal and mouse movements. The basic idea is that a person can be identified by means of gaze signals or human visual behavior. This has been shown several times~\cite{holland2011biometric,C2019,RLDIFFPRIV2020FUHL}. Since the gaze signal can only be computed with a webcam in a browser, it requires the user to activate and allow the access to the webcam. Additionally, the quality of the camera as well as the different lighting conditions influence the accuracy of the gaze signal~\cite{papoutsaki2016webgazer}. Further scientific work has already been done on the correlation of mouse movements and the human eye signal~\cite{liebling2014gaze,guo2010towards,navalpakkam2013measurement}. It has been found that when clicking or the end point of a mouse movement almost always corresponds to the eye position~\cite{liebling2014gaze,guo2010towards,navalpakkam2013measurement}. With this information, the technique of webcam based eye tracking has changed in that the mouse information is used to calibrate the eye tracker~\cite{papoutsaki2016webgazer}. Another advantage of mouse movements is that this information is freely accessible in the browser and does not have to be activated manually by the user like the camera. We show in this thesis that the mouse information is sufficient to identify a user, which is also scientifically based on the fact that visual behavior is user-specific~\cite{holland2011biometric,C2019,RLDIFFPRIV2020FUHL} and that mouse movements in the browser correlate with visual behavior~\cite{liebling2014gaze,guo2010towards,navalpakkam2013measurement}.
	
	The application of these data sources in the industrial environment is enormous. For example, it enables continuous user validation for online banking and external access to corporate networks. It would not be enough to have only the password and the user ID, one would also have to be able to emulate the behavior of the correct human user. For user-specific advertising and product placement, it is also possible to differentiate between users on a shared computer and identification of the same user on different computers.

	\section{\uppercase{Related work}}
	In this section, we discuss the state of the art regarding browser-based user identification. The first work which has dealt with browser-based user identification is \cite{mayer2009any}. It analyzed and used statistics about the browser configuration, version and installed extensions. In \cite{eckersley2010unique} the approach was proven in a larger study, and thus it was shown that the digital fingerprint can be effectively used for user identification via statistics. Further, studies \cite{alaca2016device,englehardt2016online,laperdrix2020browser,kobusinska2018big} dealt with extensions of the statistical features and their quality for user recognition. In \cite{alaca2016device,kobusinska2018big} an analysis of the stability of the individual characteristics was also carried out. \cite{laperdrix2020browser,kobusinska2018big} examined different browsers and also analyzed security settings that can prevent fingerprinting. There was also a long term study which dealt with the creation of a unique fingerprint over years \cite{gomez2018hiding}. Cross-browser fingerprinting was covered in \cite{cao2017cross} using both operating system and hardware features. A further extension of these approaches is the use of hashing algorithms to make the calculation and identification more effective \cite{gabryel2020browser}.
	
	Applications for browser-based fingerprints are described in the literature as user tracking \cite{eckersley2010unique,englehardt2016online}, abuse prevention \cite{vastel2020fp}, and authentication \cite{alaca2016device} in many contexts. For example, security companies use the fingerprint to detect bots or abnormal behavior on web pages \cite{misc17,misc18}. In \cite{vastel2020fp} it is also shown that fingerprinting can be used to easily block scripts that collect data from web pages, but the authors also show that this protection can be easily circumvented. A fingerprint for mobile devices, which was calculated on all hardware and installed software, is described in \cite{bursztein2016picasso}. This makes it possible to distinguish between the real device and a simulated environment of the same device, and thus block network traffic in case of a simulated device.
	
	Literature that deals with abuse prevention is mostly in the context of advertising or e-commerce. These concerns click fraud or credit card payments. Two new inference techniques were presented in \cite{nagaraja2019clicktok}. The first technique recognizes click patterns within an advertising network and thus can prevent click fraud. In the second technique, bait clicks are injected and resulting conspicuous patterns are detected. \cite{renjith2018detection} deals with credit card fraud. Here, cheap goods or services are sought that have never been shipped or performed. The authors use different features and machine learning algorithms to detect this type of fraud.
	
	There is also already some work in the field of deep neural networks for fraud detection. In \cite{zhang2019hoba} a deep neural network was presented, which analyses the data for similar behavior. This allows fraud cases, which are repeated and follow the same procedures, to be detected and traced. This technique also helps to protect against known fraud, because the behavior is conspicuous for the system. Several interconnected neural networks have also been used to detect intrusion into computer networks \cite{ludwig2019applying}. Here, various deep neural networks are used to monitor network communication. These networks detect patterns in the communication which do not correspond to the norm and warn early in case of a possible intrusion. Deep Boltzmann machines were used for fraud detection in biometric systems \cite{de2019deep}. For this purpose, features from the deep layers of the network were used, as these have proven to be more robust against attempts of fraud. Another use of deep neural networks for fingerprint calculation is described in \cite{salakhutdinov2009semantic}. Here, auto-encoders are used to calculate a hash of a document. Similar documents produce a similar hash. This technique can also be applied to browser statistics to obtain a fingerprint of a user.
	
	While click patterns\cite{nagaraja2019clicktok} or drawing symbols~\cite{syukri1998user} have already been used to calculate a fingerprint, there exists also work using mouse statistics for user identification~\cite{ahmed2007new}. In \cite{ahmed2007new} the recorded 22 subjects in 998 sessions with a session length of 45 seconds. Afterwards, they computed mouse statistics like average mouse velocity, click frequency, and a mouse angle histogram. Similar statistical features were used in \cite{chu2013blog} to detect if the current user is human or a bot. In \cite{kratky2018recognition} the mouse statistics are combined with keyboard statistics like average key press duration and key press latency. The same statistics are used in \cite{solano2020few} for a behavioral login application. The mouse statistics were also combined with the same statistics computed on the gaze of a person during game playing~\cite{kasprowski2018biometric}. A deep learning approach was proposed in \cite{hu2019insider} where the authors draw the mouse movements between two events (Key press) on the entire scene and feed it into a deep neural network. The disadvantage of this approach is, that it requires a lot of computational resources as well as it also leads to many misclassifications.
	
	We propose to use spatial distributions as well as click distributions. This leads to a reduced data representation which can be effectively used to identify the user. In addition, this data only requires a resource saving machine learning approach to identify the user.
	
	\section{\uppercase{Method \& Data recording}}
	The data was recorded on two PCs with two different browsers each. On each PC, six recordings per person were made, each of which had a minimum length of five minutes and could be of any length. On each computer, there were two browsers, each of which was used for three recordings. The choice of websites was limited to six and selected with buttons on the top and bottom of the web page. In addition, all sub-pages could be reached as well as other sites could be visited through internal links. This limitation was due to the need of an iframe, which allowed us to keep the recording software running constantly. We have also chosen to use a fixed selection of pages that are the same for all subjects, as it is guaranteed that the subjects receive the same stimulus. This reduces the influence of completely different pages on the gaze and mouse data. Before each recording, the eye tracker software WebGazer~\cite{papoutsaki2016webgazer} was calibrated. During the calibration, the subject had to gaze at each calibration point and click on each point two times. This information was given to WebGazer~\cite{papoutsaki2016webgazer} as calibration coordinates. After the calibration, the recording started. In total, six people performed the study, which brings the total number of images to 72 ($2 browser * 2 computers * 3 images * 6 people = 72$).
	
	The collected data is the gaze signal encoded as heatmap, the mouse movements as heatmap and the browser statistics. For the heatmap, we have quantized the data into a $10 \times 10$ grid, which is valid for both the eye movement and the mouse movement. In addition, we normalized the sum of the heatmap to one. The collected browser statistics are standard values like webdriver, webgl, header, language, device memory, etc. according to the FingerprintJS~\cite{miscJS}. To store the data online we used a local Apache server~\cite{wolfgarten2004apache} with a MySQL database~\cite{greenspan2001mysql}, to which the data was sent via Ajax~\cite{garrett2005ajax} using Javascript~\cite{goodman2007javascript}.
	
	\begin{figure}[h]
		\centering
		\includegraphics[width=0.12\textwidth]{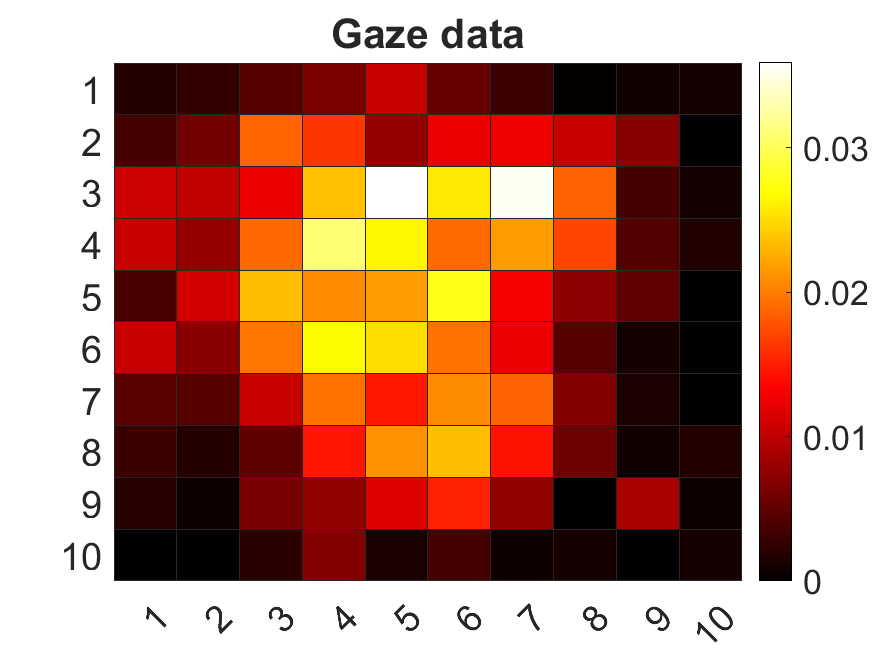}
		\includegraphics[width=0.12\textwidth]{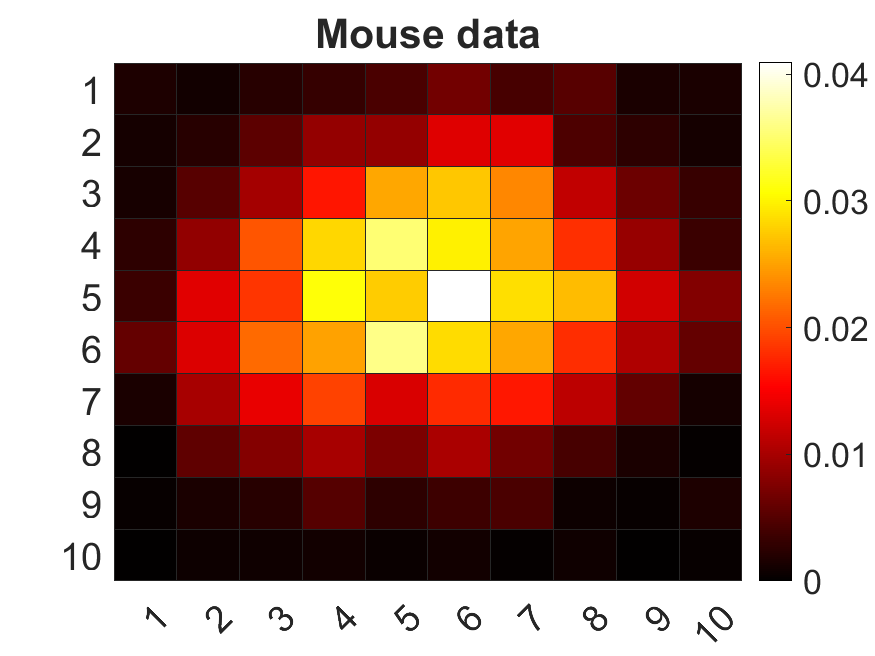}
		\includegraphics[width=0.12\textwidth]{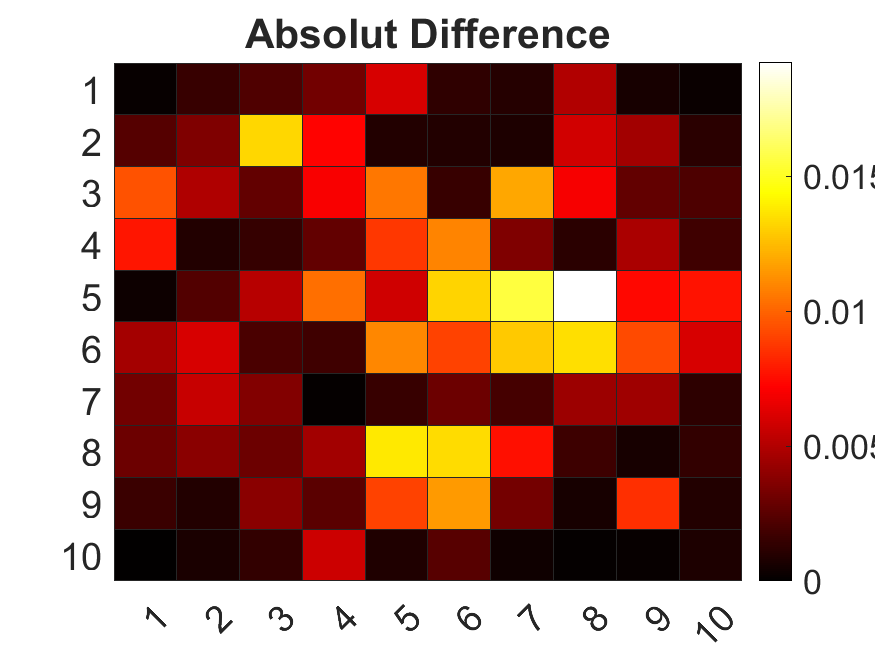}\\
		\includegraphics[width=0.12\textwidth]{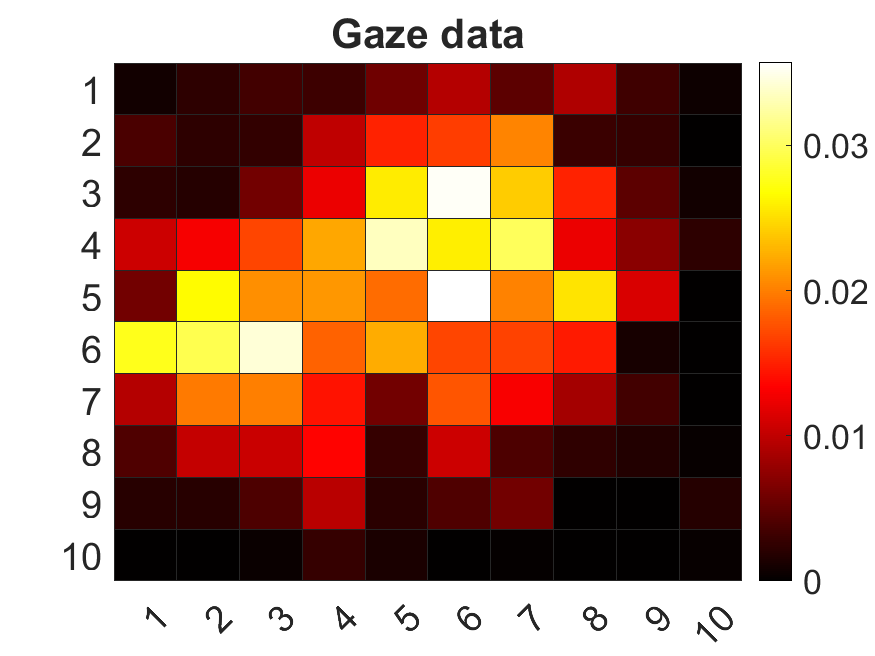}
		\includegraphics[width=0.12\textwidth]{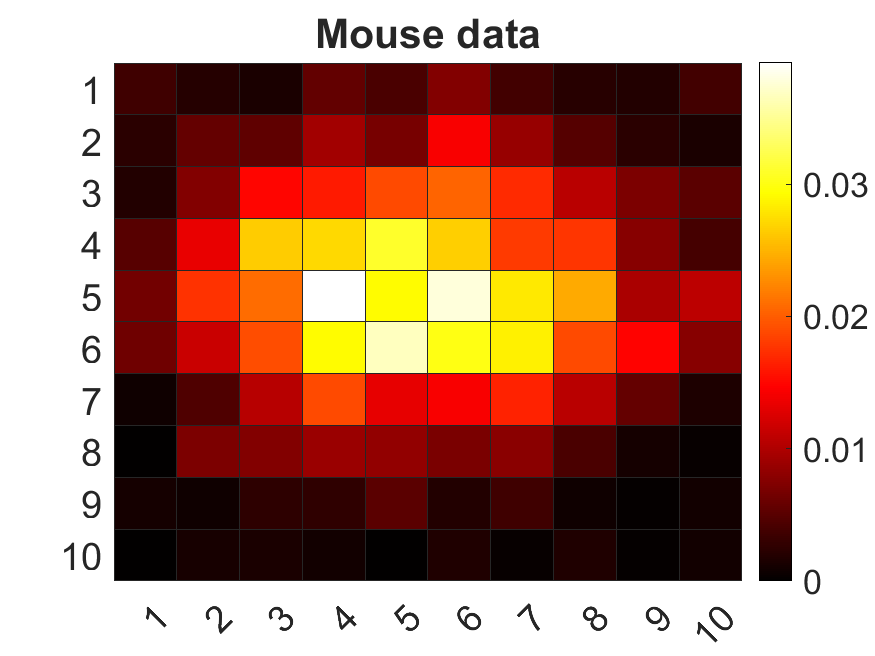}
		\includegraphics[width=0.12\textwidth]{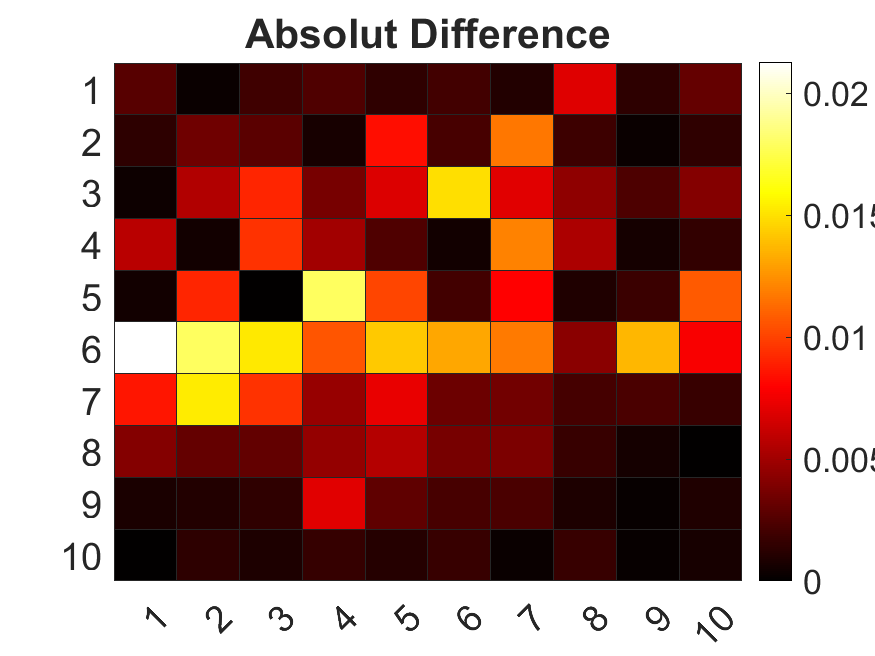}\\
		\includegraphics[width=0.12\textwidth]{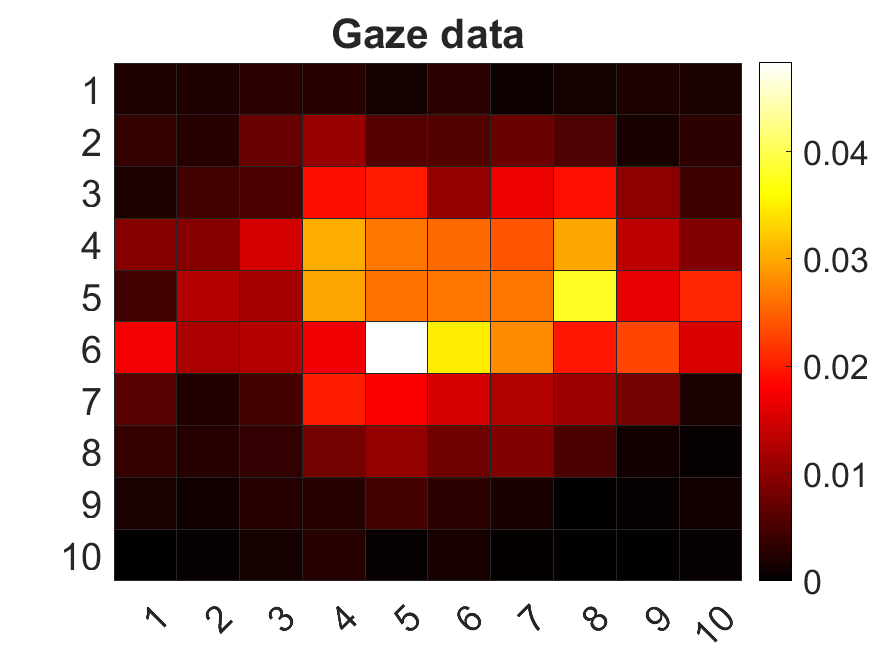}
		\includegraphics[width=0.12\textwidth]{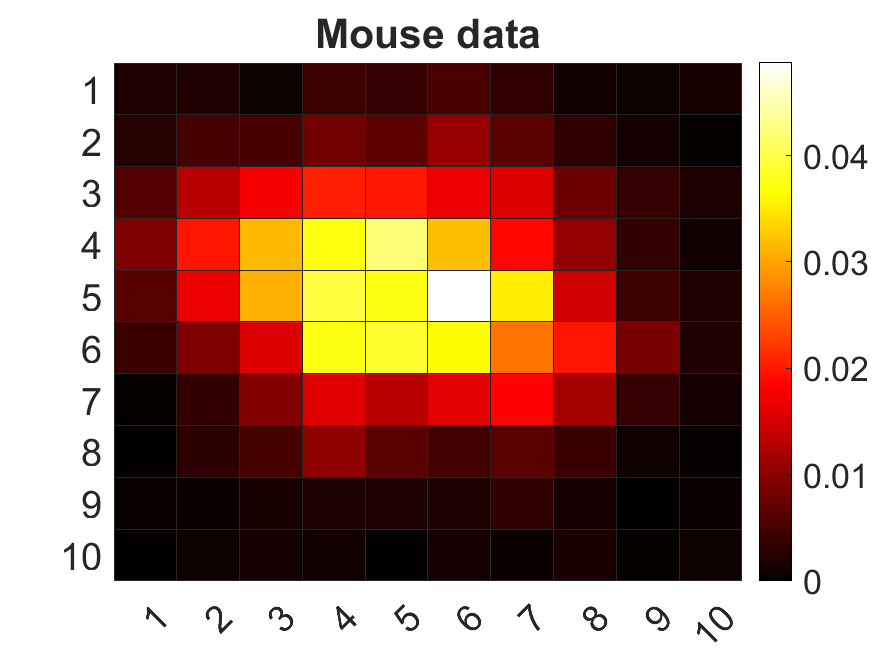}
		\includegraphics[width=0.12\textwidth]{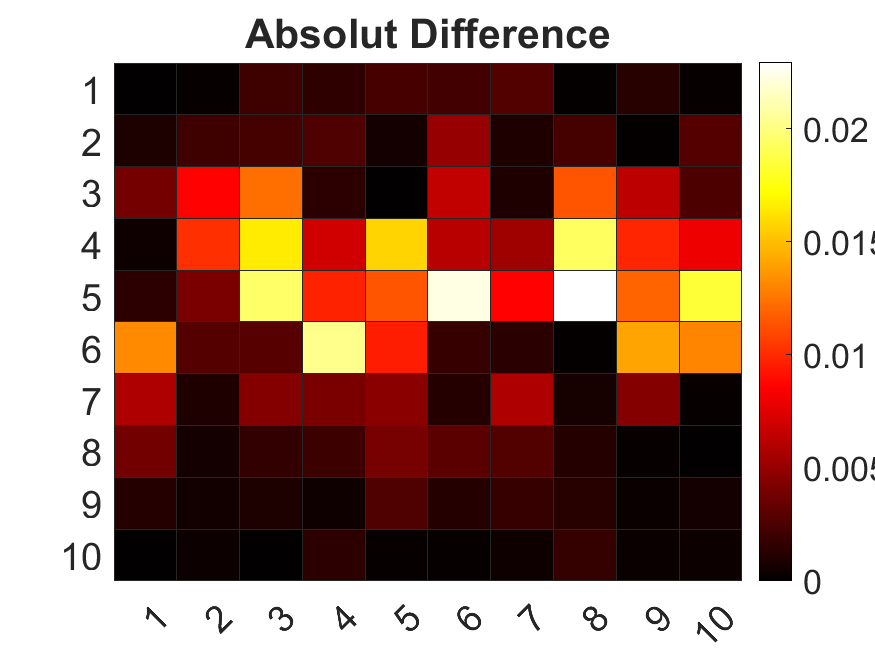}
		\caption{The gaze, mouse, and absolute difference for three users. Each row corresponds to one subject.}
		\label{fig:heat}
	\end{figure}
	Figure~\ref{fig:heat} shows the normalized gaze and mouse movement data. Each row corresponds to a separate image. Comparing the mouse and gaze data, a clear difference can be seen where both signals have the main focus relatively central. Since we used an iframe for our recordings, we could not use tabs in the browser. Scrolling was also done mainly over the mouse wheel, and the scrollbar was a bit inside the screen. In the third column in Figure~\ref{fig:heat}, which represents the absolute difference, it can be seen that the signals are clearly different. Nevertheless, the signals correlate with each other, which is of course also due to the fact that the heatmap is a quantization and is invariant to time.
	
	\begin{figure}[h]
		\centering
		\includegraphics[width=0.3\textwidth]{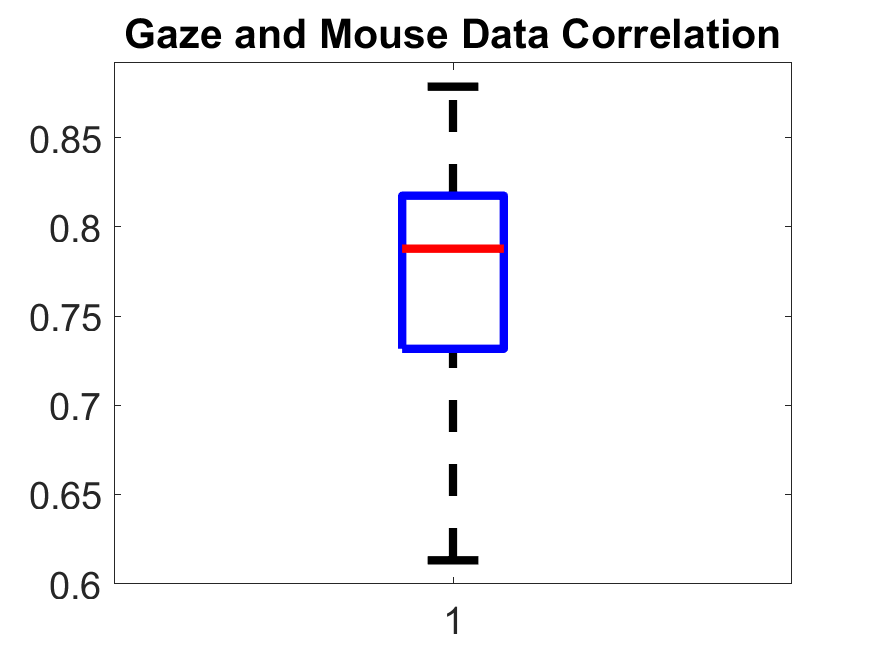}
		\caption{Correlation between the mouse and the gaze data over all samples as whisker plot.}
		\label{fig:corr}
	\end{figure}
	Figure~\ref{fig:corr} shows the distribution as a whisker plot of the correlation coefficient over all recorded data between the gaze and mouse signal. The blue box represents the 75\% confidence interval and the red line represents the median. Since all values of the correlation coefficient are above 0.5, it can be assumed that there is a very strong relation between the gaze signal and the mouse signal in our recordings. As already mentioned, this is reinforced by the heatmap quantization, the normalization, and the removal of the time dependency in the heatmap.
	
	\textbf{Larger scale study of mouse movements: } In this study, we recorded the mouse movements of 80 people. Each person made ten recordings and could determine the websites, duration, device (As long as our software was running on it) and also the activity completely freely. With this study, we would like to show that it is also possible to distinguish people on a larger scale based on their mouse movements. The age of the test persons was between 24 and 39. We recorded the mouse movements as well as the clicks with the mouse (left, right and middle mouse button as well as the mouse wheel movements upwards and downwards). The recording was done with a program running in the background so that the persons were completely free in the choice of their activity as well as in the choice of the web pages. There were also no time restrictions for the recordings, so the recording time of our data was between 4 and 25 minutes. As in the first study, we coded each entire recording into a $10 \times 10$ heatmap (quantizing the resolution into $10 \times 10$ fields) and normalized over their sum. Mouse clicks were also divided by their sum to make this usable as a distribution. In our evaluation, experiments are performed with and without the mouse click distribution.

	\section{\uppercase{Evaluation of the first small study}}
	
	\begin{figure*}[h]
		\centering
		\includegraphics[width=0.3\textwidth]{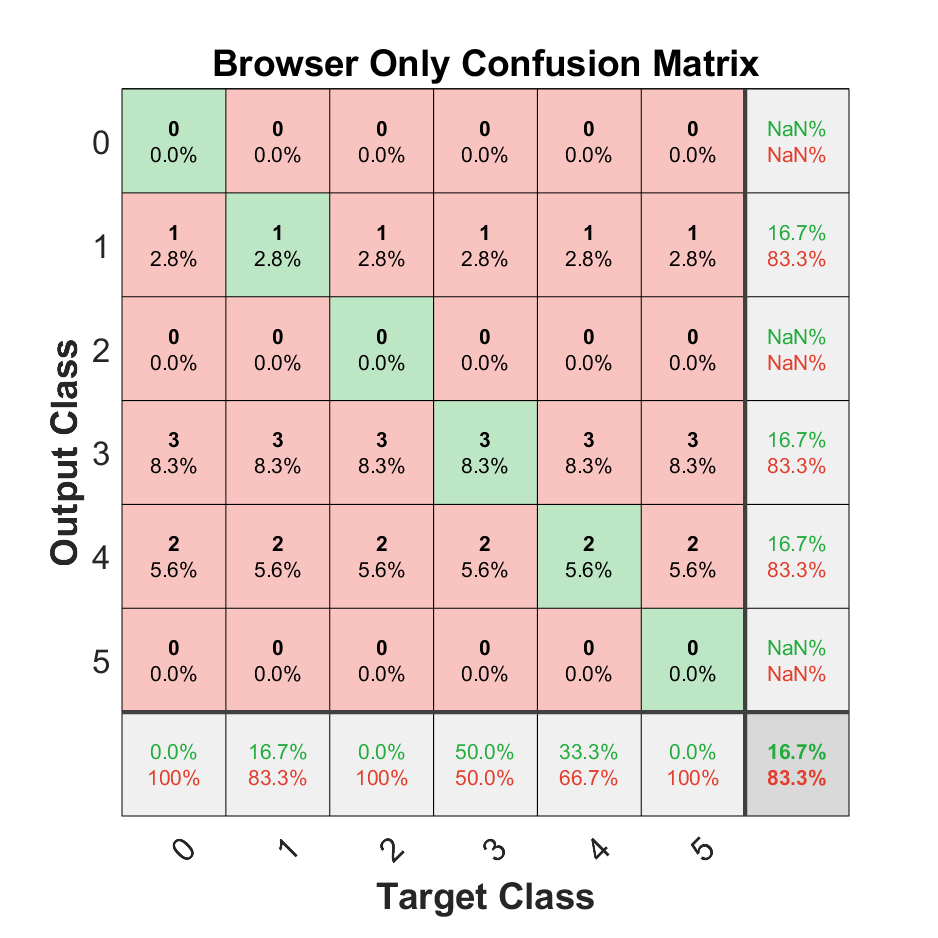}
		\includegraphics[width=0.3\textwidth]{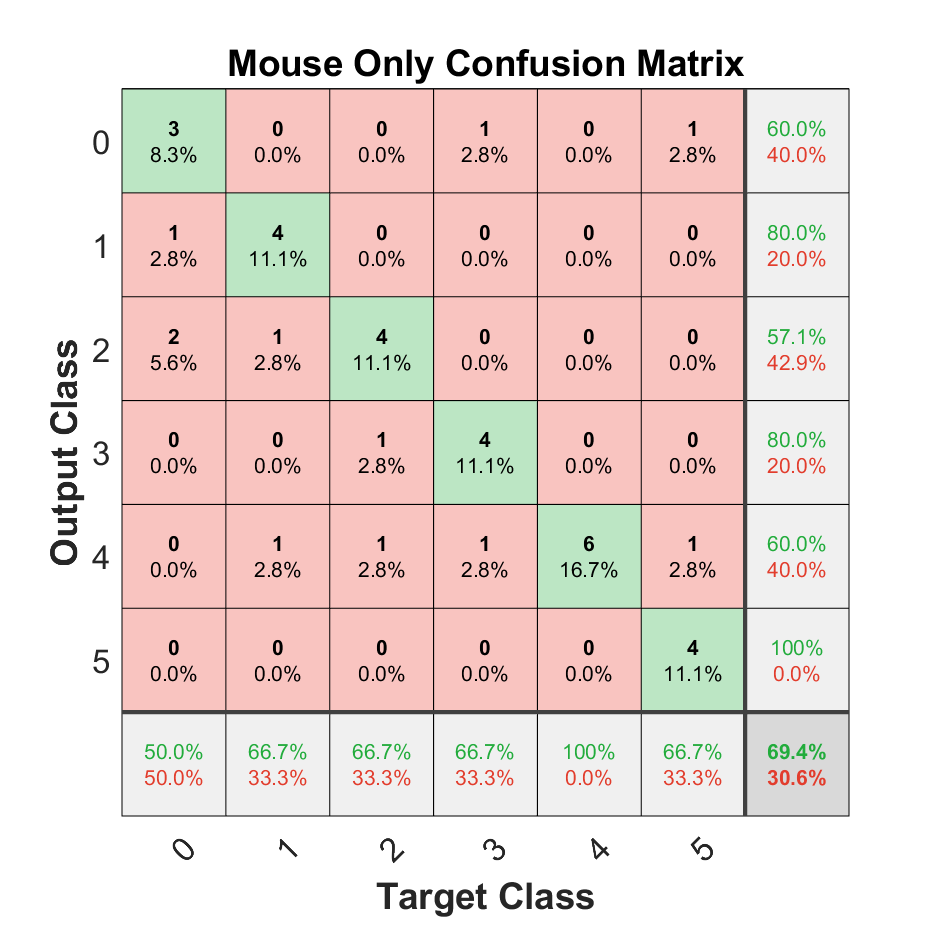}
		\includegraphics[width=0.3\textwidth]{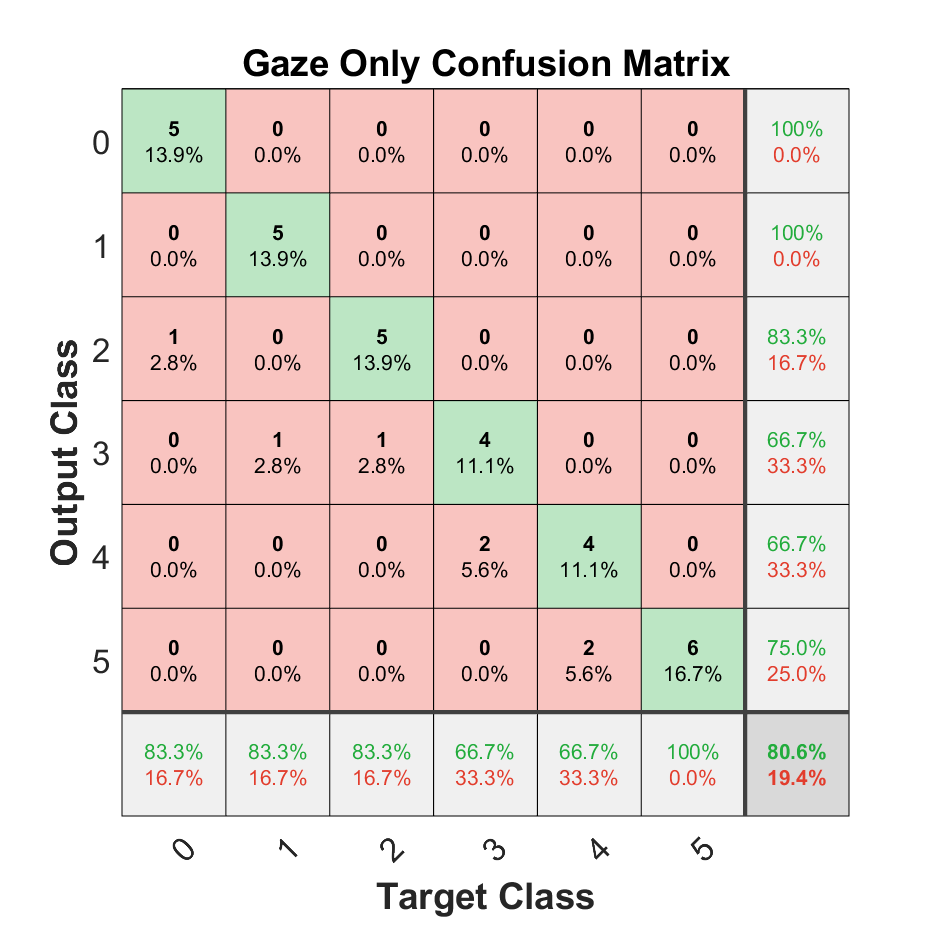}
		\caption{The confusion matrices with the browser statistics, as well as for the gaze and mouse data as input separately. On the bottom right of each confusion matrix, the overall accuracy can be seen.}
		\label{fig:eachsolo}
	\end{figure*}
	
	For all our evaluations, we performed a 50\% to 50\% split between training and validation data. We made sure that every user is present in the training data, as well as every computer and browser. This has been done because it is necessary to have seen the user at least once to recognize him. To make the comparison to the browser statistics fair, we made sure that every computer and browser is in the training data as well. As a machine learning method, we have chosen bagged decision trees with the standard parameters of Matlab 2020b. The only setting we have made is the number of trees to be trained, which we have set to 50 for all evaluations. We did not perform any data augmentation or other preprocessing. We have chosen this approach because it is the easiest to reproduce. In addition, this work is not about the best possible results, but about the proof of concept of using the mouse as well as gaze data to create a digital fingerprint. The script and data can be viewed in the supplementary material and tested together with Matlab.
	
	Figure~\ref{fig:eachsolo} shows the results for the use of the individual data sources (browser statistics, mouse, and gaze) separately. The results are displayed as a confusion matrix to view each class separately. Each confusion matrix also has the overall accuracy in the lower right corner. The matrix on the left side was only evaluated with the browser statistics as input. The overall accuracy is exactly at the chance level (16.66\%). This shows that the browser statistics, in the case of computers with multiple users, cannot be used effectively to distinguish between the different users. This is because if two users on the same computer in the same browser have the same statistical values. The middle confusion matrix in Figure~\ref{fig:eachsolo} shows the results achieved with the mouse heatmap. The overall accuracy of 69.4\% is significantly above the chance level of 16.66\%. Thus, it is clear that the mouse data contains information about the user. Furthermore, this data can be used to distinguish between users who have used the same computer and the same browser. The right matrix in Figure~\ref{fig:eachsolo} shows the results of the gaze data. These results exceed the results of the mouse data with 80.6\% accuracy. This also means that the gaze data can be used to differentiate between users on the same computer, and this even better than any other data source.
	
	\begin{figure*}[h]
		\centering
		\includegraphics[width=0.3\textwidth]{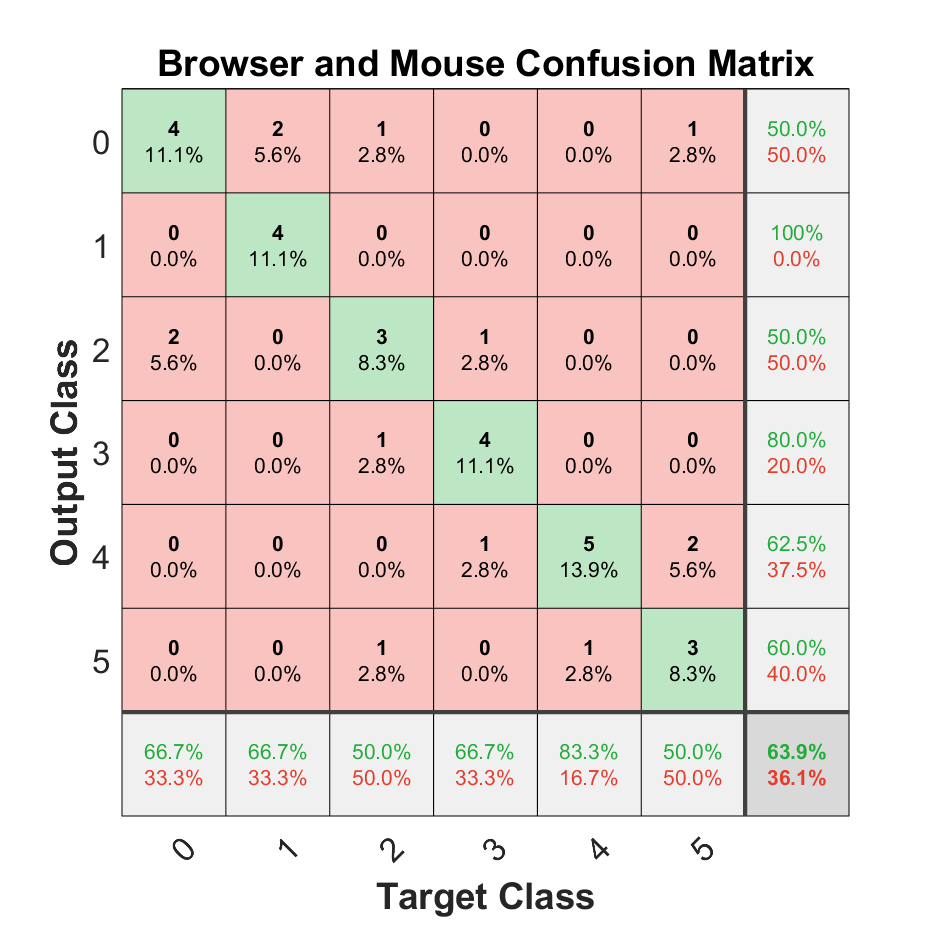}
		\includegraphics[width=0.3\textwidth]{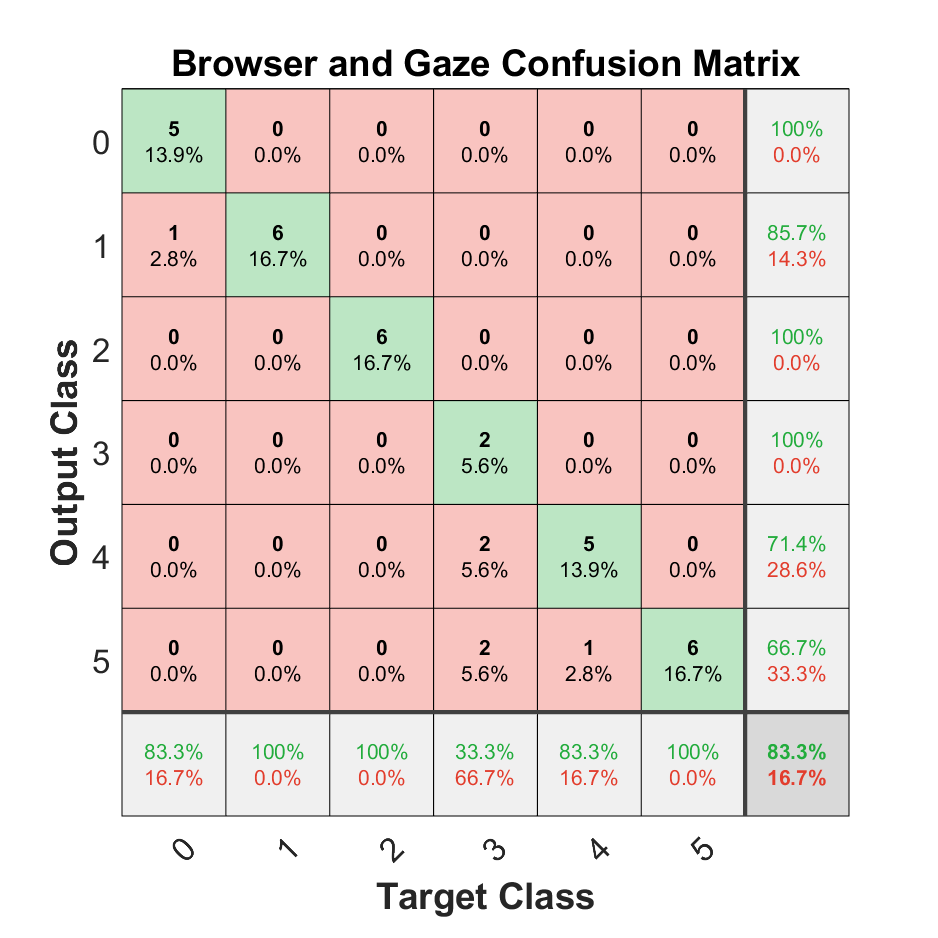}
		\includegraphics[width=0.3\textwidth]{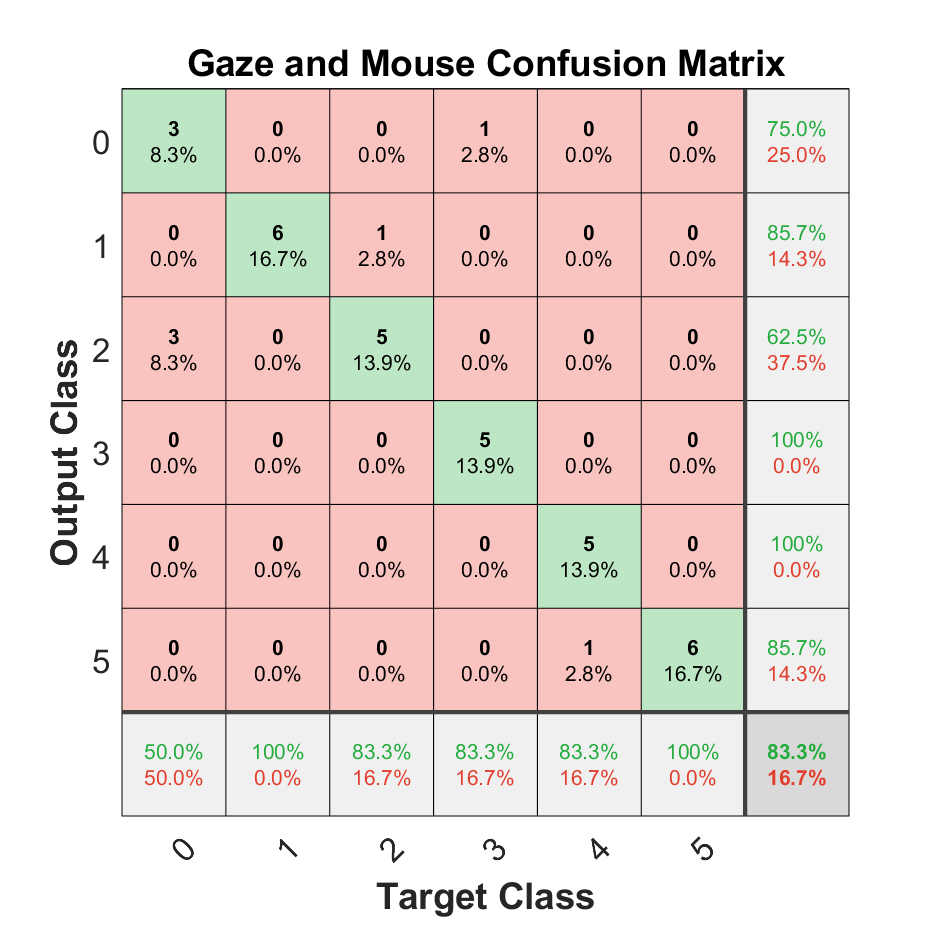}
		\caption{The confusion matrices for the combination evaluations. The left most confusion matrix is the browser and mouse data as input. In the central plot, the results for the browser and gaze data as input are shown. For the right confusion matrix, we combined the gaze and mouse heatmap. On the bottom right of each confusion matrix, the overall accuracy can be seen.}
		\label{fig:tupels}
	\end{figure*}
	
	Figure~\ref{fig:tupels} shows the results for the combinatorial use of the individual data (browser statistics, mouse, and gaze). Like the individual results, the combinatorial evaluations are also displayed as a confusion matrix. The left matrix shows the results of the combination of browser statistics and the mouse heatmap. This combination is slightly worse than using the mouse heatmap alone (69.4 to 63.9\%). This is mainly due to the fact that in our study, the users are equally distributed over all computers and browsers. Normally, that is, that every user has his own computer, except for a few, the browser statistics alone would be very effective and the combination of mouse and browser statistics would be much better. In the middle matrix of Figure~\ref{fig:tupels} the results of the combination of gaze and browser statistics are shown. Here a slight improvement can be seen (80.6\% to 83.3\%). This is not very much, but the same applies as for the combination of mouse and browser statistics. Usually the browser statistics is very effective, because most of the users have their own computer, so this combination would be much more effective. The last and right matrix in Figure~\ref{fig:tupels} shows the results of the mouse and gaze heatmaps. Here is only a small improvement to the gaze heatmaps alone (2.7\%). One reason for this is that the two heatmaps have a very similar content and therefore correlate strongly (Figure~\ref{fig:corr}). In addition, larger data volumes of significantly more than six users would have to be included in order to be able to finally evaluate this. In general, it can be assumed that this combination is not very effective.
	
	\begin{figure}[h]
		\centering
		\includegraphics[width=0.3\textwidth]{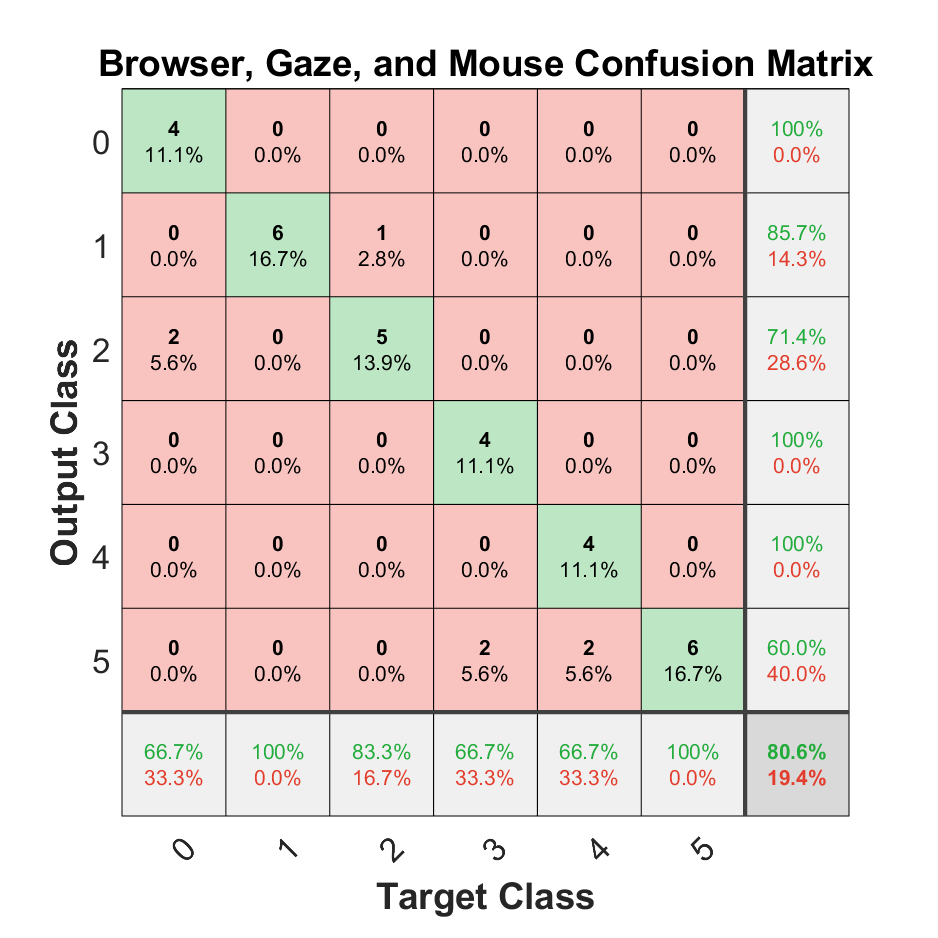}
		\caption{The confusion matrix with the combination of all data sources which are the browser statistics, the gaze heatmap, and the mouse heatmap. On the bottom right, the overall accuracy can be seen.}
		\label{fig:alltogether}
	\end{figure}
	
	Figure~\ref{fig:alltogether} shows the results for the combinatorial use of all data together (browser statistics, mouse, and gaze). As in all previous evaluations, we have used a confusion matrix. The overall result of all data as input is as good as the result when using the gaze data alone (80.6 to 80.6). The individual values of the correct and incorrect classifications in the matrix differ slightly, but the overall result shows no improvement. Of course, as for all combinatorial analysis with the browser data, the browser data usually works very well and the result is certainly much better, if there are only a few users which share a computer. Also, the combination of the gaze and mouse heatmap may not be optimal, because these data correlate too strongly. However, the gaze heatmap can be replaced with the mouse heatmap. The mouse data have the clear advantage that they can always be retrieved and do not require calibration.
	
	\section{\uppercase{Evaluation of the larger scale study}}
	
	\begin{table}[htb]
		\caption{Shows the average validation results of a 5 folds cross validation. The task for the different classifiers was the classification of the person (80 persons in the dataset) based on a heatmap or a heatmap and click distribution. We used the standard parameters of the Matlab classification learner and compared to a reimplementation of the state-of-the-art mouse statistics (Mouse stats).}
		\label{tbl:PersonClassification}
		\centering
		\setlength\tabcolsep{1pt}
		\begin{tabular}{lccc}
			Method & Heatmap & Heatmap and clicks & Mouse stats\\ \hline
			BaggedTree & 90.75\% & 90.5\% & 83.5\% \\
			Discriminant & 98.125\% & 98.375\% & 87.125\% \\
			KNNEnsem. & 94.0\% & 92.5\% & 90.0\% \\
			CubicSVM & 93.5\% & 93.625\% & 35.75\% \\
			LinearSVM & 97.0\% & 95.25\% & 86.375\% \\
			GaussSVM & 91.0\% & 88.375\% & 82.25\% \\
			QuadricSVM & 95.125\% & 95.375\% & 41.625\% \\
			FineKNN & 97.125\% & 97.75\% & 87.25\% \\
			WeightKNN & 97.25\% & 96.625\% & 89.625\% \\
		\end{tabular}
	\end{table}

	
	\begin{figure}[h]
		\centering
		\includegraphics[width=0.23\textwidth]{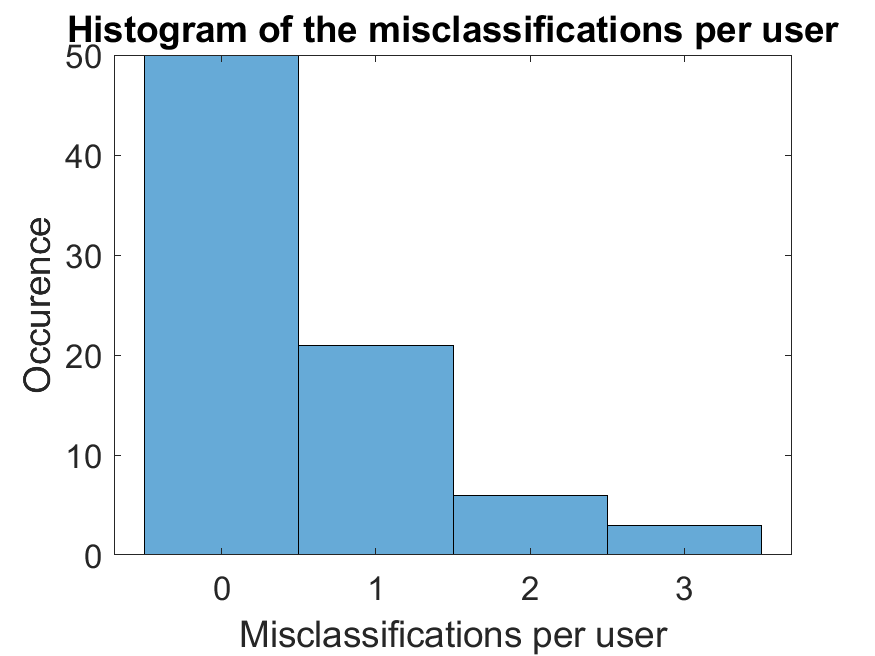}
		\includegraphics[width=0.23\textwidth]{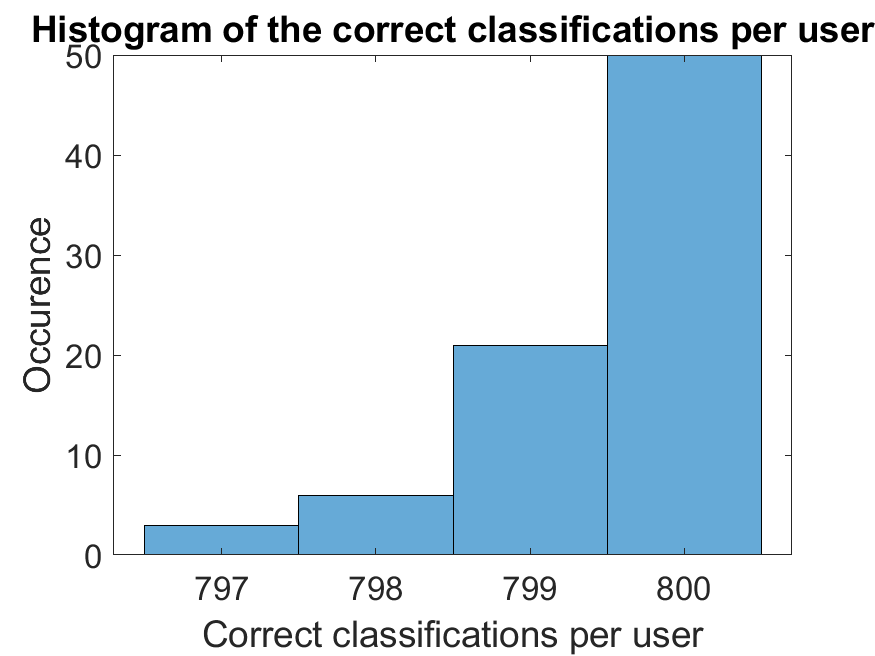}
		\caption{Histogram of the misclassification occurrence per user on the left and the histogram of correct classification occurrence per user on the right for the one vs all evaluation per subject without the click distribution (All confusion matrices for each user are in the supplementary material). This means that each user had to be distinguished from all other users based on a binary classification. We conducted this experiment since it should be closer to the usage as a security mechanism in online banking or similar, where it is about validating the user based on his behavior. For each user, we conducted a 5-fold cross validation and did not balance the dataset (Which can be seen in the confusion matrices based on the numbers of class one and two) nor used any reweighting mechanism. The results for all confusion matrices are from the Matlab 2020b FineKNN with the standard parameters as they are used in the classification learner application. The reason for this histogram evaluation is to show that the approach works for all users.}
		\label{fig:onevsall}
	\end{figure}

	\begin{figure*}[h]
		\centering
		\includegraphics[width=0.23\textwidth]{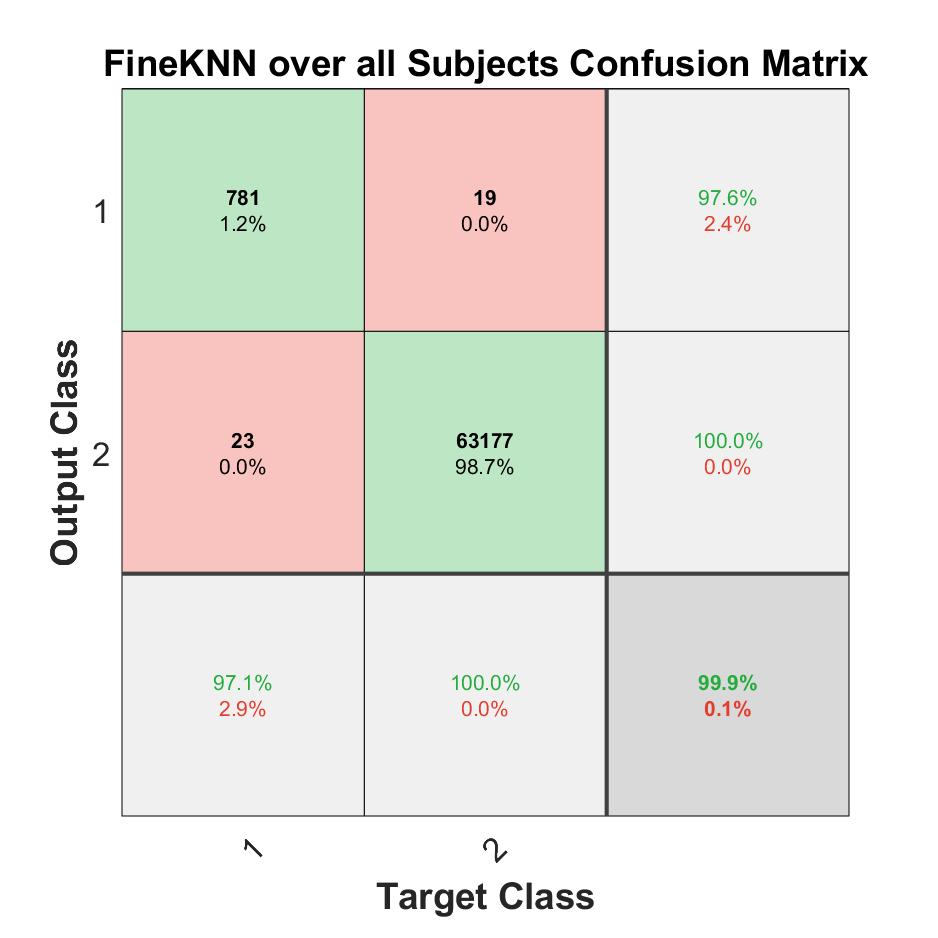}
		\includegraphics[width=0.23\textwidth]{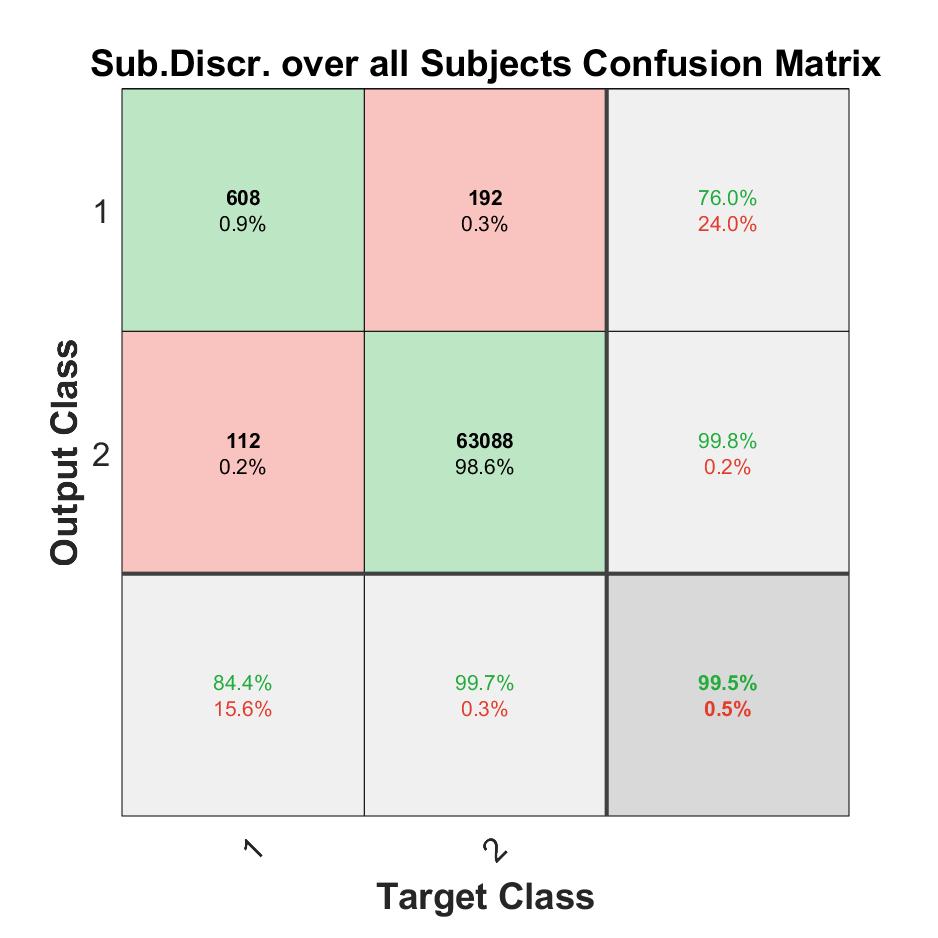}
		\includegraphics[width=0.23\textwidth]{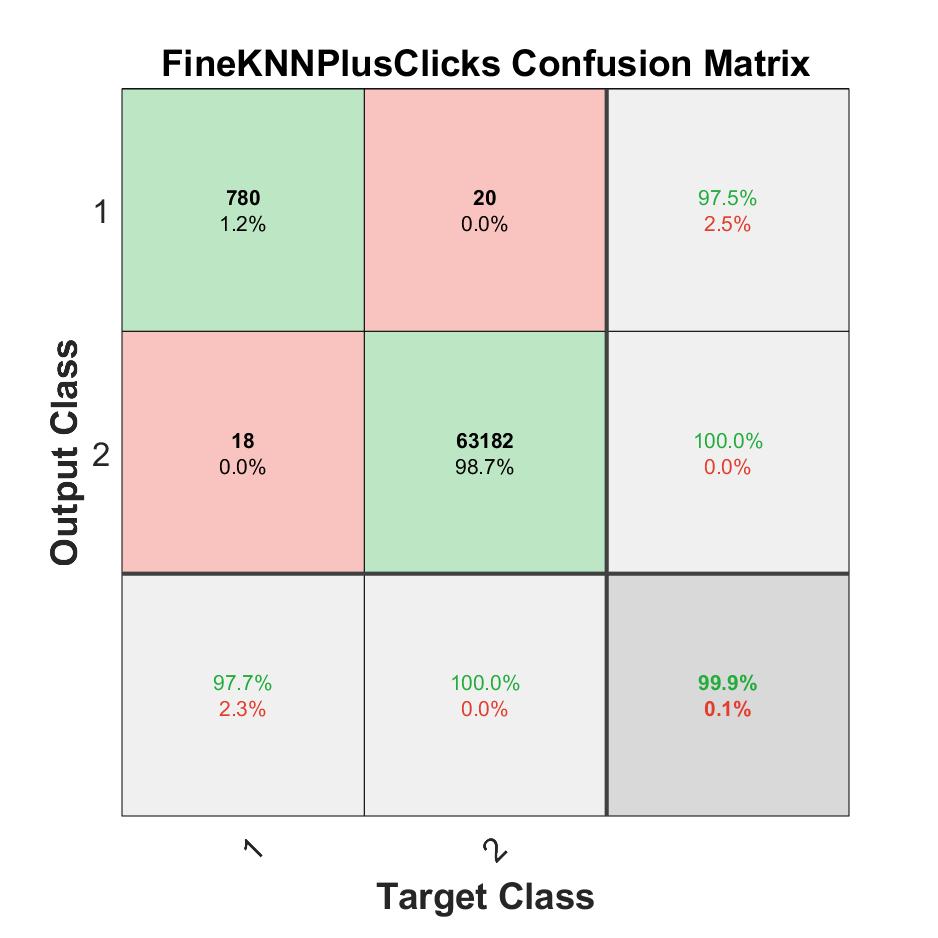}
		\includegraphics[width=0.23\textwidth]{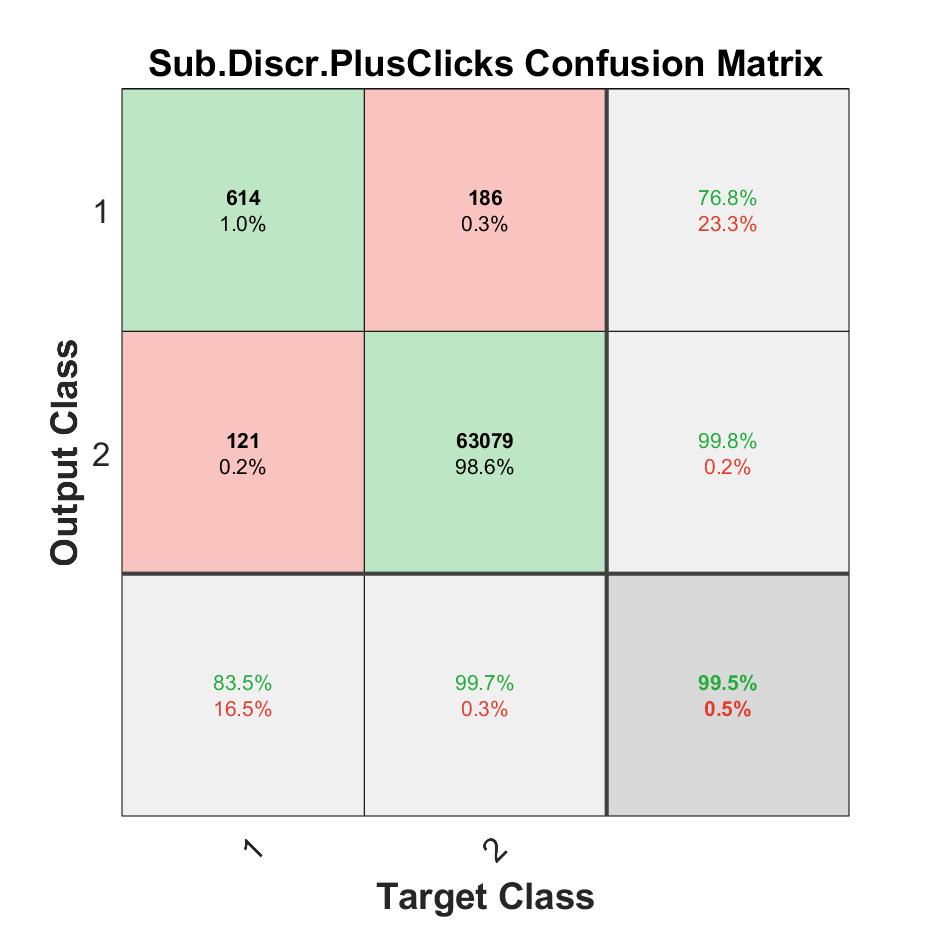}
		\caption{The confusion matrices for different machine learning methods. We conducted the same per user evaluation as in Figure~\ref{fig:onevsall} but computed the confusion matrix on all predictions together. This means we did a 5-fold cross-validation for each person vs all other persons and did not rebalance the dataset nor reweight the cost function.	This does not show that it works for each user equally good (But this was shown in Figure~\ref{fig:onevsall} already). The first two confusion matrices are with the heatmap, and the last two confusion matrices are with the heatmap in combination with the click distribution.}
		\label{fig:onevsalloverall}
	\end{figure*}
	
	Table~\ref{tbl:PersonClassification} shows the results of person classification based on the heatmap and the heatmap combined with the click distribution. The underlying data are the mouse recordings of the 80 subjects in our second study. We evaluated each method with a 5-fold cross validation and calculated the mean accuracy. As can be seen in Table~\ref{tbl:PersonClassification}, the Ensemble of subspace discriminant classifiers is the method with the best accuracy (98.125\% and 98.375\%) second is FineKNN (97.125\% and 97.75\%). For all procedures, we used the default parameters of Matlab and did not perform a grid search for the optimal parameters. This, together with the scripts and data provided, should make the results easy to reproduce. If we compare the results of the heatmap as well as the results of the heatmap with click distribution, we can clearly see that the click distribution has a rather negative effect in most cases. This is certainly due to the fact that little of an individual's behavior is reflected in the click behavior, since people are restricted in their clicks and the left mouse button and scrolling are probably the most frequently used functions. This evaluation shows that mouse movements are a very good way to detect users based on their behavior. 
	
	Since the evaluation in Table~\ref{tbl:PersonClassification} is a detection and identification of one person out of many, this form of use is rather less suitable for security-relevant scenarios such as online banking. Therefore, we have performed further experiments in which the goal is to validate one person out of many. For this, we assigned class one to a single person and class two to all others. In this way, the machine learning algorithms learn to recognize a person based on their behavior and to validate them based on their behavior. Which can basically provide an additional layer of security in online banking or other security related network services. 
	
	The evaluations per person as histograms can be seen in Figure~\ref{fig:onevsall}. Here we performed a 5-fold cross validation for each person and attempted to validate this person against all others using Matlab 2020b's FineKNN. No balancing or cost function weighting was used for training and evaluation, which means that the results can also be further improved. As can be seen in Figure~\ref{fig:onevsall} on the left, there are only rare amounts of misclassifications per user. This is especially true if the histogram of correct classifications (On the right side of Figure~\ref{fig:onevsall}) is considered.
	
	For the results in Figure~\ref{fig:onevsalloverall}, we also did a 5-fold cross validation for each individual person. However, here we computed the confusion matrix over all results to compare different machine learning methods and the heatmap in combination with the click distribution for this task. As can be seen in the confusion matrices in Figure~\ref{fig:onevsalloverall}, FineKNN is by far the best method in terms of the first row of the confusion matrix. This means that it is least likely to classify a valid person as invalid. If we compare the first 6 confusion matrices with the last 6, we see that the click distribution worsens the results, as it does for the person identification (Table~\ref{tbl:PersonClassification}).

	\section{\uppercase{Conclusion}}
	In this work, we have done a small study and analyzed both gaze and mouse data to use them for a digital fingerprint. In our evaluation, it is clearly shown that these signals can be used individually as well as in combination for fingerprinting. It also shows that in the case of computers used by multiple users, the browser statistic fails and can no longer distinguish the persons. With our data we can confirm, as in previous work, that the gaze and mouse signals are dependent on each other.

	\bibliographystyle{plain}
	\bibliography{template}

\end{document}